\documentclass[10pt,oneside]{article}

\usepackage{natbib}
\setcitestyle{authoryear} 

\usepackage[]{graphicx}
\usepackage[]{color}
\usepackage{alltt}
\usepackage{physics}
\usepackage{amsmath}
\usepackage{dsfont}
\usepackage[toc,page]{appendix}
\usepackage{subcaption}
\usepackage{array}
\usepackage{booktabs}
\usepackage{amssymb}
\usepackage{tikz}
\usetikzlibrary{bayesnet}
\usepackage[colorlinks,linkcolor=blue,anchorcolor=red,citecolor=green]{hyperref}
\usepackage[section]{placeins}
\usepackage{geometry}
\usepackage{chngcntr}
\usepackage{tikz}
\usetikzlibrary{fit,positioning,arrows,automata}
 
\DeclareMathOperator*{\argmax}{arg\,max} 
\DeclareMathOperator{\E}{\mathbb{E}}

\counterwithout{figure}{section}
\numberwithin{equation}{section}
\usepackage{amsthm}
\usepackage{enumitem}
\usepackage{ifthen}
\newcommand{\compilehidecomments}{false} 
\ifthenelse{ \equal{\compilehidecomments}{false} }{%
	\newcommand{\tk}[1]{}
	\newcommand{\es}[1]{}
	\newcommand{\us}[1]{}
	\newcommand{\yw}[1]{}
}{
	\newcommand{\tk}[1]{{\color{blue}  [Tim: #1]}}
	\newcommand{\yw}[1]{{\color{red}  [Yiliu: #1]}}
	\newcommand{\es}[1]{{\color{green} [Eric: #1]}}
	\newcommand{\us}[1]{{\color{cyan} [Uygar: #1]}}
}

\newcommand{\rev}[1]{#1}

\begin{document}

\title{Interpretable time series analysis with Gumbel dynamics}

\author{
	Yiliu Wang\thanks{Allen Institute, Lead Contact: \url{yiliu.wang@alleninstitute.org}}\thanks{University of Washington} \and
	Timothy Doyeon Kim\footnotemark[1]\footnotemark[2] \and
	Eric Shea-Brown\footnotemark[2]  \and
	Uygar S\"{u}mb\"{u}l \footnotemark[1]\footnotemark[2] 
}

\date{}
\maketitle

\begin{abstract}
		Switching dynamical systems can model complicated time series data while maintaining interpretability by inferring a finite set of dynamics primitives and explaining different portions of the observed time series with one of these primitives. However, due to the discrete nature of this set, such models struggle to capture smooth, variable-speed transitions, as well as stochastic mixtures of overlapping states (e.g., non-instantaneous state transitions), and the inferred dynamics often display spurious rapid switching on real-world datasets. Here, we propose the Gumbel Dynamical Model (GDM). First, by introducing a continuous relaxation of discrete states and a different noise model defined on the relaxed-discrete state space via the Gumbel distribution, GDM expands the set of available state dynamics, allowing the model to approximate smoother and non-stationary ground-truth dynamics more faithfully. Breaking from established literature, \rev{this new class uses a stochastic relaxed state to directly combine reusable dynamical primitives and govern observations, without a separate stochastic continuous trajectory.} Second, the relaxation makes the model fully differentiable, enabling fast and scalable training with standard gradient descent methods. 
		We validate our approach on standard simulation datasets and highlight its ability to model soft, sticky states and transitions in a stochastic setting. Furthermore, we apply our model to two real-world datasets, demonstrating its ability to infer interpretable states in stochastic time series with multiple dynamics, a setting where traditional methods often fail.
	\end{abstract}

	\section{Introduction}
	Natural behaviors give rise to complex time series data with non-stationary and nonlinear dynamics. Such dynamical phenomena are often well approximated within a temporal neighborhood by a small set of distinct, interpretable motifs~\citep{wiltschko2015mapping}. A family of dynamical system models aims to discover these discrete state transitions in an unsupervised manner. In particular, switching linear dynamical systems (SLDSs) formalize this observation by inferring a decomposition of the complex dynamics into locally linear dynamics primitives~\citep{ackerson1970state, barber2006expectation, linderman2017bayesian, glaser2020recurrent, chen2024probabilistic}. Only one of the dynamics primitives is used to describe the underlying data at any time point, which is defined as the state of the system. The model learns to switch between states to improve accuracy, enabling interpretable explanations of the observations.  \rev{However, many real-world physical and behavioral systems evolve continuously at the temporal resolution of measurement. Even when an abstract regime label changes instantaneously, the observed trajectory may pass through a finite-duration interval whose local dynamics are not well described by either endpoint primitive alone. } In such cases, interpretability of SLDS models diminishes. Moreover, switching between discrete states is prone to spurious rapid switching \rev{or requires additional transient states} under the influence of complex noise processes across multiple states, a phenomenon commonly observed in real datasets.
	
	More broadly, while desirable for interpretability, discreteness poses challenges in analyzing the physical world. One relevant manifestation is the difficulty of incorporating discrete factors into machine learning models: although gradient descent fuels spectacular successes, obtaining gradient estimates around such discrete factors is inherently problematic. The Gumbel distribution, a member of the extreme value distribution family~\citep{gumbel1935valeurs, gumbel1941return}, offers a relaxation to produce ``soft discrete'' samples, where the approximation is controlled by a temperature parameter~\citep{jang2016categorical, maddison2016concrete}. Here, we adopt this approach to propose a dynamical model that approximates switching dynamics, is trained with gradient descent, and offers interpretable characterizations even when the parameter estimates deviate substantially.
	
	The Gumbel-soft relaxation of states, the soft transition design of the dynamics, and the efficient inference algorithms together provide several advantages for analyzing complex time series. First, the model accommodates systems with mixed states and stochastic transitions. Second, the soft relaxation reduces spurious rapid switching, leading to more interpretable notions of states. Finally, the models are fast to train and generalize readily to unseen data. We validate our approach on benchmark simulations and two real-world datasets: Formula~1 race telemetry data~\citep{fastf1} and the Caltech Mouse Social Interactions dataset (CalMS21)~\citep{sun2021multi}. We observe that our implementation learns faster and produces more interpretable state estimates compared to competitive benchmarks.

\subsection{Related work}	
Our model relates to state-space models such as autoregressive HMMs (AR-HMMs) and switching linear dynamical systems (SLDSs). AR-HMMs extend HMMs with autoregressive observations~\citep{juang1985mixture, guan2016efficient}, while SLDSs decompose dynamics into linear primitives~\citep{ackerson1970state}. Extensions have improved flexibility and inference, including recurrent SLDSs~\citep{linderman2017bayesian, glaser2020recurrent}, nonlinear variants~\citep{Dong2020}, duration modeling~\citep{Ansari2021,Slupinski2025}, structured online inference~\citep{Hashempoor2026}, and smooth transitions via Gaussian processes~\citep{Hu2024}. In contrast, GDM introduces smoothness directly in the state representation, allowing overlapping and continuous state evolution without additional duration variables or boundary assumptions. \rev{GDM also directly combines dynamical primitives and removes the stochastic continuous trajectory.}

Recent studies emphasize preserving interpretability while increasing expressivity, often by decomposing time series into linear dynamical systems (LDSs). The Kalman VAE combines a variational auto-encoder with linear Gaussian state-space models and a dynamics-parameter network for time-varying weights~\citep{Fraccaro2017}. Other approaches model transitions as time-varying mixtures of LDSs or introduce hierarchical structures to encourage sparse and smooth dynamics coefficients~\citep{Mudrik2024, chen2024probabilistic}. While these improve robustness to noise, they remove the notion of discrete states and their recurrent relationships with the environment. More recently, TiDHy decomposes dynamics into orthogonal LDSs operating at different timescales~\citep{Abe2025}.

The Gumbel-Softmax distribution has been explored as a differentiable reparameterization tool in SLDS-related models. Prior work has used it to approximate discrete dynamics weights, relax categorical states for gradient-based training, or replace marginalization in variational inference~\citep{Fraccaro2017, BeckerEhmck2019, Dong2020}. \rev{GDM instead uses the stochastic relaxed state to govern observations directly, without a separate stochastic continuous trajectory (see Appendix~\ref{app:gumbels}).}

GDM is not restricted to classical SLDS parameterizations and can incorporate expressive sequence models such as RNNs. Modern architectures—including neural differential equations, S4, transformers, and recurrent deep networks~\citep{chen2018neural, gu2021efficiently, vaswani2017attention}—achieve strong performance in sequence modeling but are not designed to produce switching dynamical primitives. Integrating Gumbel-driven state dynamics with such architectures, for example by coupling attention or continuous-time models with learnable switching structure, is a promising direction for combining expressivity with interpretability.

\subsection{Summary of contributions}
Our contributions are summarized in the following points. 

$\bullet$ We propose a Gumbel-driven dynamical system  model defined over a relaxed-discrete state space that enables smooth, overlapping state transitions while retaining the interpretability of state-based dynamical primitives. The model is designed for non-stationary time series where regimes may overlap and/or evolve gradually.

$\bullet$ We formulate a fully differentiable generative model and variational posterior directly over relaxed states, enabling scalable training with standard gradient descent methods. 

$\bullet$ We design an amortized inference network for states, enabling fast inference and immediate generalization to unseen sequences {\it without re-optimizing a per-sequence latent trajectory posterior}, in contrast to many existing methods.

$\bullet$ We evaluate performance using metrics that capture both fit and quality of the inferred states. \rev{GDM improves state recovery in soft-sticky NASCAR and yields interpretable dynamical primitives in two real-world datasets.}

\section{Model formulation}
\label{sec:form}

\subsection{Gumbel-Softmax trick}

The Gumbel–Softmax trick~\cite{jang2016categorical, maddison2016concrete} provides a continuous relaxation of discrete random variables, enabling gradient-based optimization. Given logits $\pi \in \mathbb{R}^K$ for a categorical distribution, we proceed as follows. Let $G(\mu,\beta)$ denote the Gumbel distribution with location $\mu$ and scale $\beta$~\cite{gumbel1941return}. We sample $g_i \sim G(0,1)$ and form perturbed logits $\pi_i + g_i$. The maximum $\max_i\{g_i+\pi_i\}$ follows a Gumbel distribution with location $\log\sum_{j}\exp(\pi_j)$ and scale $1$, and the maximizing index follows the categorical distribution (Gumbel-Max trick):
$P(i=\argmax_j (g_j+\pi_j)) = \frac{\exp(\pi_i)}{\sum_j \exp(\pi_j)}$.

As a member of the extreme-value family, Gumbel noise amplifies differences among competing logits, sharpening the winner-take-all behavior of the Gumbel–Max construction. A continuous relaxation replaces the argmax with a tempered softmax, yielding the Gumbel-Softmax (GS) distribution
$z \sim \mathrm{softmax}\left(\frac{\pi + g}{\tau}\right)$,
where $\tau$ controls the softness. As $\tau \to 0^+$, the softmax approaches the argmax and the GS distribution recovers the categorical distribution.

The Gumbel-Max construction is invariant to shifts in the location parameter $\mu$, while the scale $\beta$ controls the spread of the noise. Sampling $g \sim G(0,\beta)$ is equivalent to sampling $\tilde{g}\sim G(0,1)$ and using
$z \sim \mathrm{softmax}\left(\frac{\pi}{\tau}+ \frac{\beta\tilde{g}}{\tau}\right)$.
Without loss of generality, we set $\beta = 1$ and write $z \sim \mathrm{GS}(\pi, \tau)$. This provides a differentiable $q(z|\phi)$ with continuous samples generated from parameter-free Gumbel noise. In practice, we set $\tau$ to a moderate value to ensure stable gradient flow while accounting for uncertainty in state transitions. Due to the extreme-value behavior of the Gumbel distribution, GS samples remain close to one-hot under moderate temperatures, preserving the semantics of discrete states while enabling smooth optimization. In the Gumbel dynamical model introduced next, \rev{Gumbel noise represents unobserved competition among primitives, and observation noise captures residual variation after their dynamics are specified.} Additional background is provided in Appendix~\ref{app:bg}.

\begin{figure}
	\centering
	\begin{tikzpicture}[
		node distance=1.0cm and 1.3cm,
		latent/.style={circle, draw, minimum size=0.8cm, text width=0.9cm, align=center, inner sep=0pt},
		obs/.style={circle, draw, fill=gray!20, minimum size=0.8cm, text width=0.9cm, align=center, inner sep=0pt},
		]
		
		\node[latent] (zm1) {$z_{t-1}$};
		\node[latent, right=of zm1] (z0) {$z_t$};
		\node[latent, right=of z0] (zp1) {$z_{t+1}$};
		
		\node[obs, below=of zm1] (ym1) {$y_{t-1}$};
		\node[obs, below=of z0] (y0) {$y_t$};
		\node[obs, below=of zp1] (yp1) {$y_{t+1}$};
		
		\edge {zm1} {z0};
		\edge {z0} {zp1};
		\edge {zm1} {ym1};
		\edge {z0} {y0};
		\edge {zp1} {yp1};
		\edge {ym1} {y0};
		\edge {y0} {yp1};
		\edge[dashed] {ym1} {z0};
		\edge[dashed] {y0} {zp1};
		\plate[inner sep=0.2cm, xshift=0.1cm, yshift=0.1cm] 
		{time1} {(zm1)(z0)(zp1)(ym1)(y0)(yp1)} {$t$};
		
		\begin{scope}[xshift=7.5cm]
			\node[latent] (zm1) {$z_{t-1}$};
			\node[latent, right=of zm1] (z0) {$z_t$};
			\node[latent, right=of z0] (zp1) {$z_{t+1}$};
			
			\node[latent, below=of zm1] (xm1) {$x_{t-1}$};
			\node[latent, below=of z0] (x0) {$x_t$};
			\node[latent, below=of zp1] (xp1) {$x_{t+1}$};
			
			\node[obs, below=of xm1] (ym1) {$y_{t-1}$};
			\node[obs, below=of x0] (y0) {$y_t$};
			\node[obs, below=of xp1] (yp1) {$y_{t+1}$};
			
			\edge {zm1} {z0};
			\edge {z0} {zp1};
			
			\edge {zm1} {xm1};
			\edge {z0} {x0};
			\edge {zp1} {xp1};
			\edge {xm1} {z0};
			\edge {x0} {zp1};
			
			\edge {xm1} {ym1};
			\edge {x0} {y0};
			\edge {xp1} {yp1};
			\edge {xm1} {x0};
			\edge {x0} {xp1};
			\plate[inner sep=0.2cm, xshift=0.1cm, yshift=0.1cm]
			{time2} {(zm1)(z0)(zp1)(xm1)(x0)(xp1)(ym1)(y0)(yp1)} {$t$};
			
		\end{scope}
	\end{tikzpicture} 
	\caption{\textbf{Graphical model representation of two systems.} Left: 2-level GDM. Right: 3-level Mixture SLDS. Dashed lines denote dependencies that can be removed \rev{for equivalence, subject to noise assumptions  (Appendix~\ref{sec:equiv}).}}
	\label{fig:graph}
\end{figure}

\subsection{Gumbel dynamical model}
We propose a new dynamic switching model to accommodate continuous Gumbel-Softmax state samples, the Gumbel Dynamical Model (GDM):
\begin{align}
	\label{eqn:gdm}
	z_1 \sim \mathrm{GS}(\pi_1, \tau), \qquad 
	& z_t \mid z_{t-1}, y_{t-1} \sim \mathrm{GS}(\pi_t, \tau), 
	\quad \pi_t = f_\theta(z_{t-1}, F y_{t-1}),  & t \geq 2, \\ \nonumber
	y_1 \mid z_1 \sim \mathcal{N}(z_1 \cdot \mu, R), \qquad 
	& y_t \mid y_{t-1}, z_t \sim \mathcal{N}\!\big(\sum_k z_{t,k} (S_{k} F y_{t-1} + b_{k}), R_t\big),  & t \geq 2.
\end{align}

Here, $\pi_1$ is a learnable prior over states, $\mu$ is an observation prior, 
$S_k \in \mathbb{R}^{N\times D}$ captures state-dependent dynamics in the projected observation space, 
$F \in \mathbb{R}^{D\times N}$ projects observations to a low-dimensional latent space, 
and $R_t$ models the observation covariance. 
Importantly, $f_\theta$ can be any feed-forward network parameterized by $\theta$. 
As a simple and interpretable case, $f_\theta$ can take a linear recurrent form 
$f_\theta(z_{t-1}, F y_{t-1}) = T F y_{t-1} + r$, where $T$ is a learnable $K\times D$ transition matrix 
and $r$ is a bias vector. 
To explicitly encourage persistence, a sticky variant mixes the logits with the previous soft state: 
$\pi_t = (1-\gamma)(T F y_{t-1}+r) + \gamma\, z_{t-1}$.

The Markov-1 assumption in the GDM can be relaxed to incorporate longer history. In this case, we parametrize the transition logits with an RNN: 
let $h_t$ be the hidden state updated as $h_t = g(h_{t-1}, F y_{t-1})$ where $g$ is a recurrent architecture such as a GRU. We then define the transition logits as $\pi_t = \textrm{FNN}(z_{t-1}, h_t)$. While the state dynamics become non-linear, the soft states $z_t$ still correspond to interpretable dynamical motifs, preserving the interpretability of the model. 

The GDM can be a highly flexible modeling tool. At the temperature limit, the linear GDM can be seen as an input-driven ARHMM with structured, low-dimensional dynamics, which makes exact inference tractable without requiring variational methods. At the other extreme, the model has a natural extension as a deep GDM, in which multiple layers of GDM are composed to increase expressivity: $h_t^{(l)} = \sum_k z_{t,k}^{l}A_k^{(l)}h_{t-1}^{(l)} + W^{(l)}h_t^{(l-1)}$ where $h_t^{(1)} = x_t$ denotes the latent projected observation (i.e., $x_t = \mathbb{E}[F y_{t} \mid z_{\leq t}]$; see Appendix~\ref{sec:equiv} for details). The standard GDM corresponds to the single-layer case. This formulation highlights that GDM spans a spectrum from interpretable linear switching models to more expressive hierarchical dynamical systems. 

\subsection{Relationship to SLDS}

GDM is closely related to SLDS, but differs in how stochasticity is modeled. In particular, GDM places stochasticity in the relaxed discrete states via Gumbel noise, rather than in a Gaussian latent trajectory. This leads to a 2-level formulation where state dynamics directly govern observations without introducing additional latent noise sources.

A related 3-level formulation can be derived by introducing latent projected observations \rev{$x_t = \mathbb{E}[F y_{t} \mid z_{\leq t}]$ for $t \geq 1$, where the expectation is taken conditional on all past states.} Replacing $F y_{t-1}$ in the GDM with $x_{t-1}$ yields a two-level GDM system, which is equivalent to
\begin{align}
	\label{eqn:3-level}
	z_1 \sim \mathrm{GS}(\pi_1, \tau), \quad 
	& z_t \mid z_{t-1}, x_{t-1} \sim \mathrm{GS}(\pi_t, \tau), \;\; \pi_t = f(z_{t-1}, x_{t-1}), \quad t \geq 2, \\ \nonumber
	x_1 = z_1 \cdot \mu, \quad 
	& x_t \mid x_{t-1}, z_t = \sum_k z_{t,k}\,(A_{k} x_{t-1}+ c_{k}), \quad t \geq 2,  \nonumber
\end{align} 
and $y_t \mid x_t \sim \mathcal{N}(C x_t, Q_t)$, for $t \geq 1$. Here, the continuous latent trajectory $x_t$ at time $t$ is determined by a mixture of dynamics over the soft states $z_t$. Importantly, $x_t$ is deterministic given $z$, and is introduced to facilitate interpretation. At each time $t$, $x_t$ can be viewed as the expected projection of $y_t$.  Uncertainty in the system is thus captured solely by the Gumbel noise on $z$ and the Gaussian noise on $y$. Figure~\ref{fig:graph} illustrates the graphical models of the two systems, highlighting their relationships and dependencies. \rev{Note that exact distributional equivalence requires allowing temporally correlated noise, while under homogeneous, conditionally independent noise, deterministic dynamics agree but noise processes differ}. A proof of system equivalence is provided in Appendix~\ref{sec:proof}. 

More generally, one could allow additional noise in the latent trajectory $x$ by introducing state-dependent covariances.  This results in a mixture version of the standard recurrent SLDS with Gumbel state dynamics.  Although more expressive in principle, the trajectory dynamics $x$ and the state dynamics $z$ compete to explain the data, and inference becomes more expensive as a flexible posterior is required to capture their intricate dependencies.  For completeness, we discuss variational inference for this 3-level mixture model in Appendix \ref{app:vi3}.

 \subsection{Discussion on interpretability}
Interpretability of GDM arises from two complementary aspects. First, GDM is a linear dynamical model, meaning that the evolution of latent dynamics is governed by a collection of linear dynamical primitives. Each primitive corresponds to a time-invariant system with well-defined analytical properties, making the learned dynamics directly analyzable. Second, GDM improves state estimation by removing the trade-off between stochasticity in the continuous latent trajectory $x$ and stochasticity in the switching state $z$, thereby enhancing interpretability in practice. To quantify interpretability, we introduce the following metric that measures the quality of inferred states at the state level.

\textbf{Inferred State Accuracy.} 
Let $\{ \zeta_t \}_{t=1}^T$, \rev{$\zeta_t \in \{1, \dots, C\}$}, denote the ground-truth (or expert-labeled) discrete states, \rev{where $C$ is the number of annotation categories,} and let $\{ z_t \}_{t=1}^T$, with $z_t \in \Delta^{K-1}$, denote the inferred states, where $\Delta^{K-1}$ is the $(K\!-\!1)$-simplex. In particular, discrete inferred states are represented as one-hot vectors in $\Delta^{K-1}$. We train a $k$-nearest neighbor (k-NN) classifier \rev{$f_{\mathrm{KNN}} : \Delta^{K-1} \to \{1, \dots, C\}$} on the training set by mapping inferred states $z_t$ to ground-truth $\zeta_t$. For test data $\mathcal{D}_{\mathrm{test}}$, predictions are obtained as
$\hat{\zeta}_t = f_{\mathrm{KNN}}(z_t), 
\quad t \in \mathcal{D}_{\mathrm{test}}$. The \emph{Inferred State Accuracy} is then defined as
$\text{Acc}_{\text{state}} 
= \frac{1}{|\mathcal{D}_{\mathrm{test}}|} 
\sum_{t \in \mathcal{D}_{\mathrm{test}}} 
\mathbf{1}\!\left[ \hat{\zeta}_t = \zeta_t \right]
$.

When the underlying ground truth $\zeta_t$ is manually obtained by human annotators, \rev{$\text{Acc}_{\text{state}}$ measures annotation agreement, providing external validation of the learned dynamical primitives.} This metric depends on the representation of latent states. Models with discrete one-hot states map to a finite set of points in the simplex, limiting their ability to represent a larger number of ground-truth classes. In contrast, continuous state representations can induce a finer partition of the simplex, enabling more flexible alignment with annotations.
We therefore interpret $\text{Acc}_{\text{state}}$ as reflecting both state interpretability and representational flexibility. Importantly, it serves only as a proxy: interpretability in GDM primarily stems from the recovery of stable, reusable dynamical primitives, while agreement with annotations provides external validation of their relevance.

Finally, we note that interpretability is also related to identifiability, which asks whether the latent representation can be recovered from observed data in a statistically well-posed way. For GDM, while the projection matrix $F$ and dynamic matrices $S_k$ are identifiable only up to an invertible linear transformation, the remaining parameters are identifiable up to permutations~\citep{BalsellsRodas2023} in the limiting case $\tau\to 0$. \rev{Note that finite-temperature identifiability remains open (Appendix~\ref{app:identifiability}).}

\section{Model Inference}
\label{sec:algo}
Due to the continuous nature of states, GDM can be trained using standard gradient descent. 
To infer GDM, we use BBVI~\citep{ranganath2014black} with Gumbel-Softmax samples (GS-BBVI): we define a variational distribution $q(z)$, sample soft states $z$ from $q(z)$, and compute unbiased samples of the ELBO gradient.  

{\bf ELBO.} The ELBO for the GDM can be written as follows:
\begin{align*}
	&\log p_\theta(y_{1:T}) \geq \E_{q(z)}\left[\log p(y,z) - \log q(z)\right]\\
	= &\mathbb{E}_{q(z)} \left[\log p(z_1) +\log p(y_1|z_1)+\sum_{t=2}^{T} \log p(y_t | y_{t-1}, z_t) + \sum_{t=2}^{T} \log p(z_t | z_{t-1},y_{t-1})   \right]- \mathbb{E}_{q(z)} \left[  \log q(z_{1:T}) \right]
\end{align*}

\subsection{Variational posteriors}
We approximate the posterior over latent states with an amortized variational distribution $q_\phi(z_{1:T} \mid y_{1:T})$, parameterized by a neural network that maps observations 
to Gumbel-Softmax logits. Specifically,
$q_\phi(z_{1:T} \mid y_{1:T}) = q_\phi(z_1 \mid y_{1:T})\prod_{t=2}^T q_\phi(z_t \mid z_{<t}, y_{1:T})$,
where each $z_t$ is a continuous Gumbel-Softmax random variable with logit $\pi'_t$ and temperature $\tau$. 

Since $z_{1:T}$ are continuous, we do not define a discrete transition matrix as in the categorical case. Instead, we directly parameterize the logits $\pi'_{1:T}$ via an inference network $g_{\phi}(y_{1:T})$, which shares a similar structure to the transition network $f_\theta$. In principle, $g_\phi$ can be more expressive than $f_\theta$, improving posterior approximation and training speed. However, an overly expressive $g_\phi$ may compensate for limitations of $f_\theta$, yielding good fit but less interpretable dynamics. For this reason, we keep their structures aligned.

Concretely, if $f_\theta$ is linear, $g_\phi$ can also be linear, e.g., $\pi'_t = W y_t + b$. A sticky component depending on $z_{t-1}$ can be added to encourage persistence, $\pi'_t = W y_t + B z_{t-1} + b$, with $z_1$ drawn from a Gumbel-Softmax distribution parameterized by learnable prior logits $\pi'_1$. This yields a Markovian factorization
$q(z_{1:T}|y_{1:T}) = q(z_1 \mid y_1)\,\prod_{t=2}^T q(z_t \mid z_{t-1}, y_t)$. If $f_\theta$ is recurrent, we instead parameterize $g_\phi$ with a bidirectional RNN or Transformer so that $\pi'_t$ depends on both past and future observations, e.g., letting $e_{1:T} = \text{BiGRU}(y_{1:T})$, we set $\pi'_t = \textrm{FNN}(z_{t-1}, e_t)$. This yields a more expressive posterior that leverages temporal context to infer $z_t$.

Thanks to the Gumbel-Softmax reparameterization, we can sample $q(z)$ sequentially in a differentiable manner. The temperature $\tau$ controls the smoothness of state transitions. Empirically, GDM is largely insensitive to $\tau$ over a broad range $\tau \in (0.5,1)$, likely due to the extreme-value nature of the Gumbel distribution, which concentrates samples near simplex corners. In practice, fixing $\tau \approx 1$ provides stable optimization and interpretable state recovery. Annealing from a higher temperature can further improve robustness, although accurate state recovery ultimately depends on learning the transition parameters rather than relying on low temperature alone.

Importantly, amortized variational inference with differentiable $q(z)$ is a key advantage of GDM. The inference network learns a reusable mapping from observations to state logits, allowing new data to be processed directly without re-optimization. This contrasts with many existing models, which
typically require re-optimizing a posterior for the latent trajectory on each new dataset.

\subsection{Smoothing and prediction} 
Once the variational posterior and model parameters are trained, the inferred system can be used for smoothing current observations, evaluating quality of fit, predicting future steps, and generating new observations.  Given a time series $y_1,\ldots, y_T$ of length $T$, we first obtain samples $z_1,\ldots, z_T$ from the variational posterior. Smoothed observations $\hat{y}_1,\ldots, \hat{y}_T$ are then computed based on the sampled states and past observations, providing a measure of reconstruction quality.  

To predict future steps, we apply the learned transition model to generate next-step states $\hat{z}_2,\ldots, \hat{z}_T$ from the sampled states $z_1,\ldots, z_{T-1}$ and preceding observations $y_1,\ldots, y_{T-1}$. These predicted states are then used to generate corresponding next-step observations $\hat{y}_2,\ldots, \hat{y}_T$. The predicted observations can be recursively fed back into the transition model, enabling multi-step-ahead predictions.  We note that an analogous procedure applies to the 3-level mixture formulation. Instead of propagating predicted observations, we propagate the inferred latent trajectory $\hat{x}_2,\ldots, \hat{x}_T$, which serves as input to the state transition function.  

Due to stochasticity in the latent states, multi-step predictions naturally yield uncertainty envelopes that reflect transition ambiguity. A $k$-step-ahead prediction for a series $y_1,\ldots, y_T$ is equivalent to producing $k$ future observations at each of the $T$ possible starting points. Because of the injected Gumbel noise in the latent states $z$, prediction trajectories may diverge after only a few steps, particularly at higher temperatures $\tau$. These divergent possibilities form a prediction envelope, whose width increases at points of greater transition uncertainty. This widening envelope corresponds naturally to the unpredictability observed in real-world dynamical systems. We will further illustrate this concept via simulation examples in section \ref{sec:numeric}.

\section{Experiments}
\label{sec:numeric}
We evaluate GDM on both simulated data and two real-world datasets. We begin with a standard, deterministic simulated example, then introduce soft, sticky, and stochastic transitions. We further evaluate the model on two real-world datasets that feature multiple and highly unpredictable transitions. To assess model performance, we use two complementary metrics. At the observation level, we compute the coefficient of determination $R^2$ between the smoothed and true observations, which quantifies the quality of fit. At the state level, we use the inferred state accuracy, introduced in Section~\ref{sec:form}, that measures the agreement between inferred and ground-truth states. We also report per-iteration wall-clock time across methods and datasets in Appendix~\ref{app:setup}. Our code is available at: \url{https://github.com/yiliuw/GDM/}.

\subsection{From deterministic to uncertain: Synthetic NASCAR dataset}
The synthetic NASCAR dataset~\citep{linderman2017bayesian} emulates cars going around a track, with four states: two for straightaways and two for semicircular turns. In the standard setting, the state evolution is nearly deterministic given the previous trajectory, resulting in a nearly fixed path from a given starting point (see Appendix~\ref{app:nascar} for details).

\begin{figure}[htbp]
	\centering
	\includegraphics[width=\textwidth]{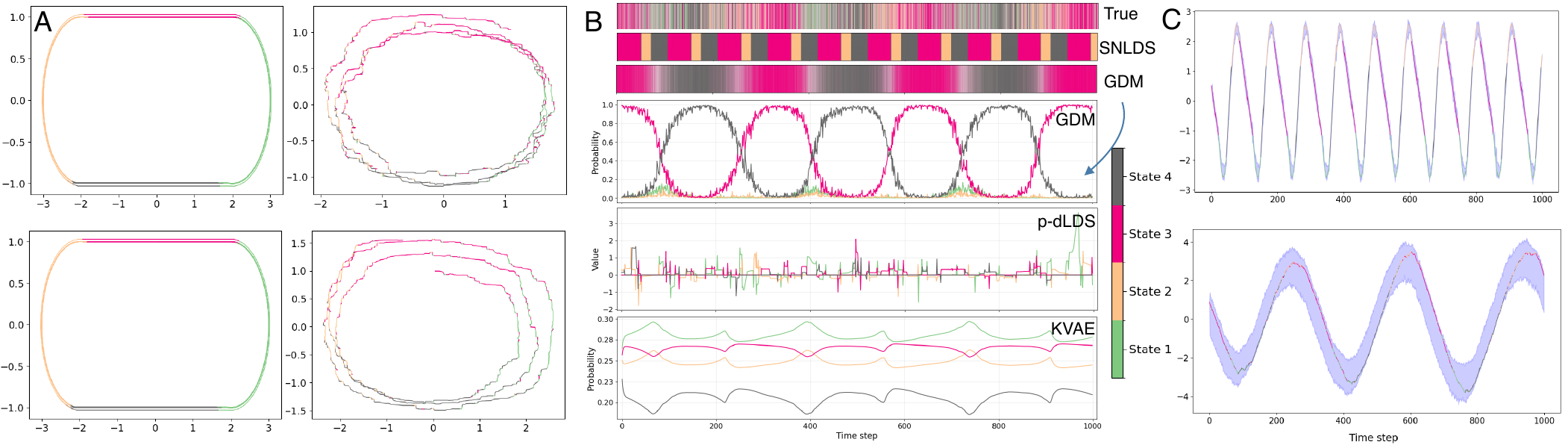}
	\caption{A. Standard and soft sticky NASCAR tracks. Two trials are generated with the same set of parameters ($T=1000$, and $K=4$ for both cases, $c=0.02$ and $\gamma=0.25$ for the soft-sticky case only). Compared to the standard NASCAR, soft sticky NASCAR introduces greater transition uncertainty. B. True states and exemplar inferred states from GDM ($\tau=0.99$), SNLDS, p-dLDS and KVAE. For each method, the panel shows the inferred state responsibilities over time: probabilities for GDM and KVAE, and scalar-valued dynamic coefficients for p-dLDS. Colored curves correspond to different dynamic primitives. Results shown are representative of 10 random seeds; full variability is reported in Table~\ref{tab:nascar}. C. Inferred 1-step-ahead prediction ranges for the first dimension of NASCAR observations. The top panel shows the standard model, and the bottom panel shows the soft sticky model, with a much wider uncertainty range. Shaded regions indicate ±3 standard deviations around the predicted mean, estimated from 100 Monte Carlo samples per model. } 
	\label{fig:nascar}
\end{figure}

We also consider a more realistic variant with soft, stochastic transitions by replacing the recurrent relationship in Eqn.~(\ref{eqn:trans}) with a soft sticky form:
\begin{equation}
	z_{t}|x_{t-1}\sim\mathrm{GS}(\pi_t, \tau), \quad 
	\pi_t = c(1-\gamma)(Sx_{t-1}+s)+\gamma z_{t-1}, \quad t\geq 2.
\end{equation}
Here, $c$ controls transition softness and $\gamma$ controls stickiness. Decreasing $c$, increasing $\gamma$, or raising $\tau$ leads to more stochastic transitions (Figure~\ref{fig:nascar}A).

We benchmark against ARHMM, SLDS, rSLDS variants, p-dLDS, KVAE, and SNLDS. All models achieve near-perfect train $R^2$ ($\approx$0.99) on both datasets. However, in the soft-sticky setting, most baselines require re-optimizing variational posteriors to achieve good test $R^2$ (otherwise the test $R^2$ is simply 0.8, i.e., the difference between the top and bottom trials), whereas GDM generalizes directly due to amortized inference (Section~\ref{sec:algo}). KVAE and SNLDS also generalize without retraining, but rely on additional architectures or marginalization strategies that complicate state interpretability. Full training setups are provided in Appendix~\ref{app:setup}.

Figure~\ref{fig:nascar}B compares inferred states across methods. GDM recovers the dominant states and represents the remaining structure through soft combinations. In contrast, baseline models struggle to capture stable and interpretable state structure in this regime: SLDS, rSLDS, and SNLDS often exhibit state collapse; KVAE produces noisy mixtures; and ARHMM and p-dLDS fail to reproduce the correct oscillatory patterns. These limitations are consistent with known challenges of discrete-switching models under soft and stochastic transitions.

Table~\ref{tab:nascar} reports the average state quality measured by mapping inferred states to hard-thresholded ground-truth states on the test trial. For the standard NASCAR data, rSLDS with recurrent only transitions achieves the top performance, while our model outperforms all the benchmarks in the soft sticky NASCAR case. Our model treats the observations as inherently stochastic, as discussed in section \ref{sec:algo}. While this uncertainty aspect is not advantageous in the standard NASCAR case, it allows the model to generalize better in the soft-sticky NASCAR case. \rev{This contrast highlights GDM's advantage when transitions are gradual, stochastic, or overlapping.} Indeed, GDM correctly identifies that the soft sticky case exhibits greater uncertainty. This is illustrated by the one-step-ahead prediction envelopes in Figure~\ref{fig:nascar}C. While most one-step-ahead observations fall inside the envelopes for both cases, the envelope is clearly wider in the soft sticky case.
\begin{table}[t]
	\centering
	\resizebox{\columnwidth}{!}{%
		\setlength{\tabcolsep}{2.5pt}
	\begin{tabular}{lcccccccc}
		\toprule
		& ARHMM& SLDS (S) & rSLDS (S) & rSLDS (R) &  p-dLDS &KVAE (M) &SNLDS (C) & GDM \\
		\midrule
		Standard & 70 ± 13& 82 ± 13 & 76 ± 10 & 96 ± 06& 74 ± 01& 67 ± 16&64 ± 02&88 ± 10 \\
		Soft-sticky &32 ± 02& 32 ± 02 & 33 ± 01 & 43 ± 09 & 34 ± 02&50 ± 11&45 ± 05& \textbf{70} ± 03 \\
		\bottomrule
	\end{tabular}
}
\caption{Comparison of $100\times$ inferred state accuracy on the standard and soft-sticky NASCAR datasets (mean $\pm$ SD). ``S'' denotes sticky variants, ``R'' denotes recurrent-only variants, ``M'' denotes the KVAE with MLP encoders, and ``C'' denotes the SNLDS with collapsed variational inference. Each model is trained and evaluated 10 times with different random seeds. \rev{On soft-sticky NASCAR, GDM remains the strongest method under top-3 and best-run reporting conventions. In particular, its lowest-scoring run (0.640) exceeds every baseline’s best run (highest: 0.589 for rSLDS(R)).}}

\label{tab:nascar}
\end{table}

\rev{As further ablations, we compare against the categorical limiting case and test whether the gains come from richer state representations only.  We compare with an input-driven recurrent ARHMM with the same $K$, observation family, split, and evaluation. This is the direct categorical limit of linear GDM: both use the preceding observation to determine next-state logits, but recurrent ARHMM selects one categorical regime whereas GDM permits stochastic relaxed combinations. The gain from $0.33$ to $0.70$ state accuracy supports the value of stochastic mixtures over categorical switching. GDM also has the highest mean Hungarian accuracy, ARI, and NMI after argmax-hardening inferred and ground-truth states (Table~\ref{tab:capacity-controls}); the advantage thus persists without continuous features for the classifier.  }
	
\begin{table}[htbp]
	\centering
	\rev{\begin{tabular}{lcccc}
			\toprule
			Model & State accuracy & Hungarian & ARI & NMI \\
			\midrule
			GDM & $0.70\pm0.03$ & $0.55\pm0.13$ & $0.30\pm0.13$ & $0.32\pm0.09$ \\
			Recurrent ARHMM & $0.33\pm0.02$ & $0.35\pm0.04$ & $0.05\pm0.07$ & $0.07\pm0.06$ \\
			rSLDS(R) & $0.43\pm0.09$ & $0.49\pm0.06$ & $0.15\pm0.08$ & $0.21\pm0.07$ \\
			\bottomrule
	\end{tabular}}
	\caption{\rev{Soft-sticky NASCAR controls (mean $\pm$ SD). State accuracy uses the original $k$-NN metric; the remaining metrics use argmax-hardened states.}}
	\label{tab:capacity-controls}
\end{table}

\subsection{From simple states to more states: F1 dataset}
The NASCAR dataset described above represents a simple track with four synthetic segments. Next, we consider a more complex and realistic example: the Formula One (F1) World Championship racetracks. We use the FastF1 package to retrieve telemetry data from past F1 sessions, including trajectory, lap times, and corner counts.

\begin{figure}[!htb]
\centering
\includegraphics[width=\textwidth]{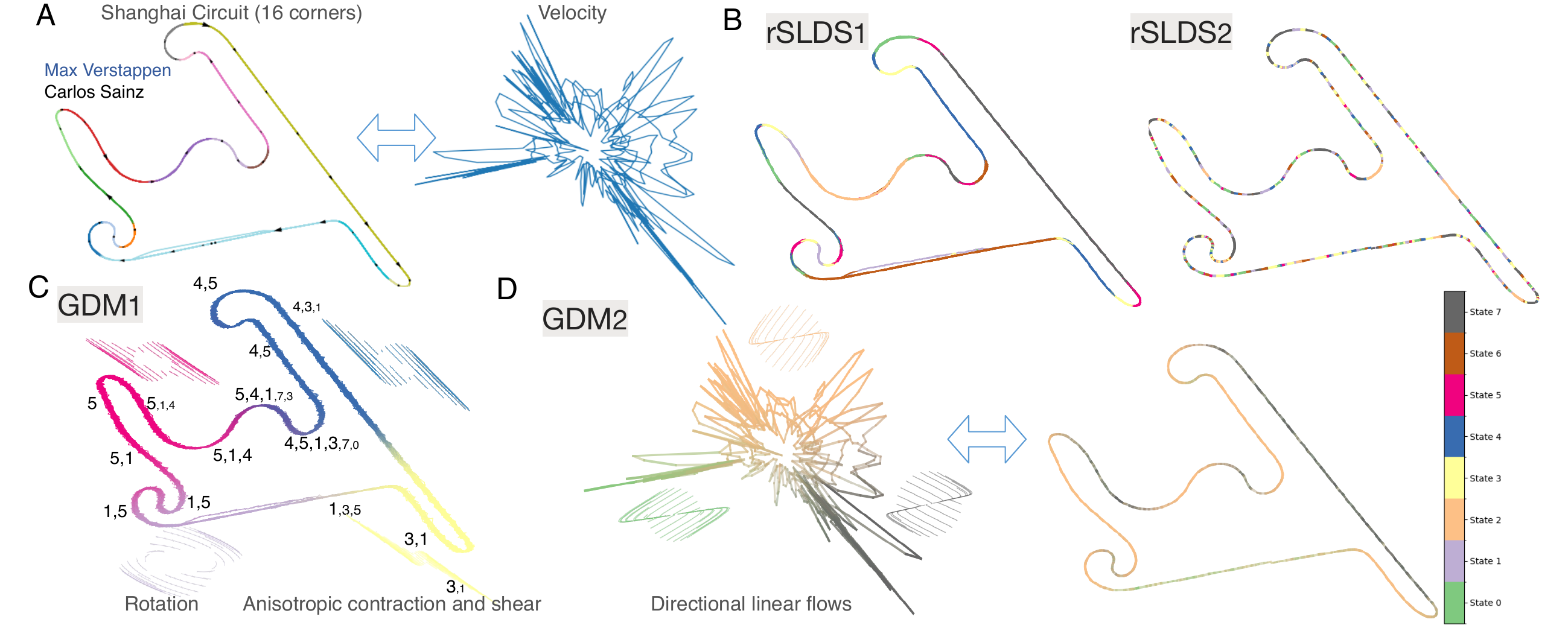}
\caption{A. Two versions of the F1 dataset with different input observations. \textit{Position-based}: experiments on individual tracks (Suzuka and Shanghai), with training and testing performed on the same track using 1st (blue) and 5th-place (black) trajectories. \textit{Velocity-based}: experiments across 24 tracks, with training on 23 tracks and testing on a held-out track (Shanghai). B-D. Example inferred trajectories for rSLDS and GDM on the Shanghai Circuit (representative of 5 random seeds). Results for other models are shown in Appendix~\ref{app:usage}. For GDM$_1$, we annotate each segment with state IDs that exceed 1\% weight in at least 20\% of the time steps within the corresponding expert-labeled segment. State IDs are ordered by presence ratio, and marker sizes reflect their weights (see Appendix~\ref{app:usage} for state usage). For GDM$_2$, the velocity-based setting yields consistent latent ``velocity states'' shared across circuits, from which the track trajectory is reconstructed. For each dominant mode, we visualize the phase plot of the corresponding dynamical system for interpretability, initialized from 24 points and evolved for 20 steps.}
\label{fig:f1}
\end{figure}

We consider two complementary versions of the F1 experiment, corresponding to different notions of ``state''. The first uses positional data as input, where we study individual F1 circuits separately, with training and testing performed on the same track using 1st- and 5th-place trajectories. In this setting, ground-truth states are defined by officially annotated corners, so similar maneuvers at different spatial locations are treated as distinct states. As shown in Figure~\ref{fig:f1}A, the Shanghai International Circuit has 16 corners. The second version uses velocity as input and is trained across multiple race tracks, thereby emphasizing shared dynamical structure across circuits rather than spatial segmentation. In this setting, no ground-truth labels are available, and inferred states are expected to reflect common local dynamics (e.g., acceleration patterns).

For the position-based experiment, we compare inferred trajectories on the Shanghai circuit at $K=8$, selected via held-out likelihood. GDM identifies a small number of dominant states and represents the remaining structure through soft combinations of these primitives. As shown in Figure~\ref{fig:f1}C, the learned dynamics consist of anisotropic shear and rotation: the former collapses trajectories onto low-dimensional directions (dominant eigen-directions), producing piecewise-linear segments and corner-like transitions under regime switching, while the latter induces smooth turning behavior. In contrast, ARHMM and rSLDS exhibit more frequent switching and less stable state assignments, particularly around corner transitions (Figure~\ref{fig:f1}B and Figure~\ref{fig:extra}). SNLDS mitigates rapid switching relative to SLDS due to its increased flexibility, but this comes at the cost of slightly distorted latent trajectories. Moreover, these models rely on discrete switching and have difficulty capturing smooth or overlapping regimes, while incurring substantially higher computational cost (see Appendix~\ref{app:setup}). Notably, all models achieve comparable training and testing fit, but discrete-switching models often do so at the expense of state quality when $K$ is small. Appendix~\ref{app:usage} provides additional results corresponding to various choices of 
$K$.

For the velocity-based experiment, we train on 23 tracks from the 2022 season and test on a held-out track (Shanghai). Due to the computational cost of nonlinear baselines, we compare GDM with linear baselines in this setting. Here, GDM groups together segments with similar local dynamics (e.g., curvature or acceleration), even when they occur at different spatial locations, whereas both ARHMM and rSLDS fail to recover such shared structure. As shown in Figure~\ref{fig:f1}D, eigenanalysis of the dynamic matrices reveals distinct linear flow structures with different directional contraction and coupling, which manifest as direction- and curvature-dependent acceleration behaviors across states. These experiments highlight that the notion of ``state'' depends on the representation: position-based inputs emphasize spatial segmentation, while velocity-based inputs reveal reusable dynamical motifs.

\subsection{Uncertainty and multiple states: CalMS21 dataset}
\begin{figure}[htbp]
	\centering
	\includegraphics[width=\textwidth]{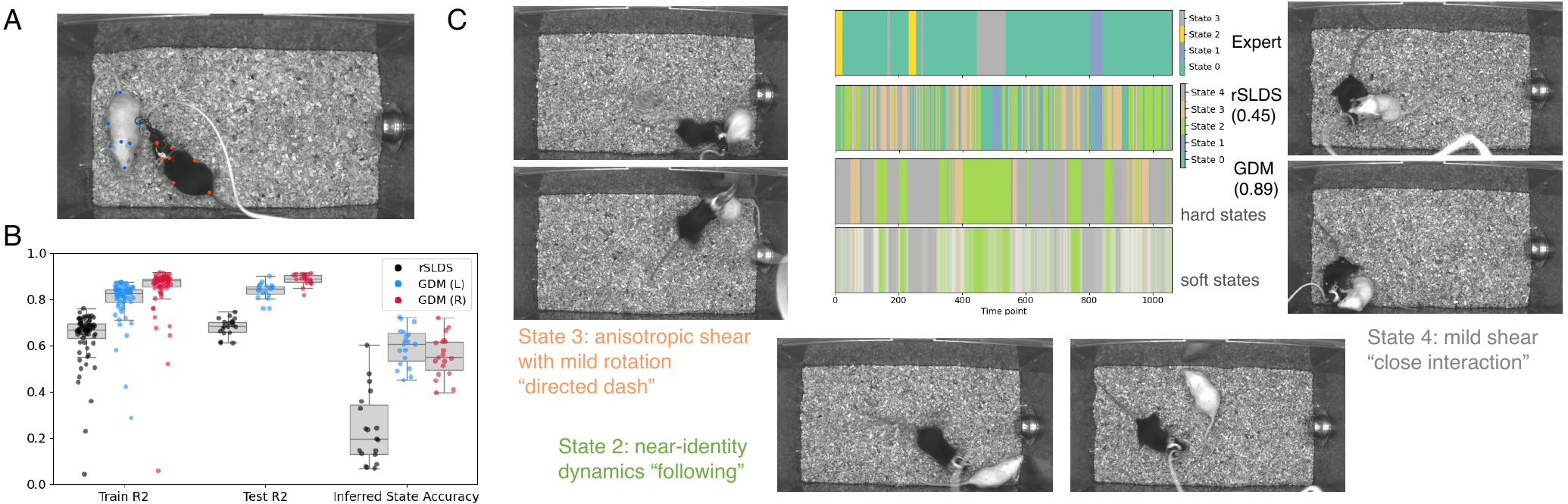}
	\caption{A. Example frame from the CalMS21 data. Seven anatomically defined keypoints are labeled on the body of each mouse. Expert annotations refer to behaviors initiated by the black mouse. B. Comparison of train $R^2$, test $R^2$, and inferred state accuracy between our models and the benchmark model rSLDS. ``L'' denotes the linear sticky GDM, and ``R'' denotes the RNN-based GDM. 
		C. Expert-labeled states and inferred states from GDM and rSLDS for Trial 34 (the shortest trial containing all states). Two representative frames are shown for each GDM dominant mode.}
	\label{fig:calms}
\end{figure}

Finally, we apply our method to study mouse social behavior using the first task in the open CalMS21 dataset~\cite{sun2021multi}. The dataset consists of location data for two mice interacting in a cage from multiple trials (89 trials, split into 70 train and 19 test) over 5 years. Each mouse is labeled with 7 keypoints, corresponding to the nose, ears, base of neck, hips, and tail (Figure~\ref{fig:calms}A). As there are 14 keypoints with x, y values per frame, the observation dimension is 28. Importantly, this dataset is expert labeled. All frames in the 89 trials are manually labelled by one expert for four distinct social behaviors (attack, investigation, mount and other).

\rev{Human annotations provide one label per frame, while the underlying motion may vary continuously or combine multiple local dynamics.} We train our models on the 70 training trials and test them on the 19 test trials, fixing the state dimension as $K=5$, as selected by GDM based on \rev{held-out ELBO without expert labels}. \rev{Here, $K$ counts dynamical primitives, not annotation categories.} In practice, initializing with a larger $K$ can be beneficial, as redundant states typically become auxiliary and contribute minimally, consistent with Occam’s razor.

Figure~\ref{fig:calms} summarizes the performance of our models (linear-sticky and RNN-based GDMs) alongside the rSLDS benchmark. Both GDM variants achieve higher training and testing accuracy than rSLDS for nearly all trials in this dataset. Consistent with our findings on the F1 dataset, our models also yield substantially higher inferred-state accuracy across all test trials. A key observation emerges when comparing the two GDM variants. The RNN-based GDM model achieves the highest observation-level accuracy for both training and test sets, reflecting the benefit of incorporating nonlinear recurrent functions into both the generative model and the variational posterior. However, its inferred-state accuracy is lower than that of the linear–sticky version. This underlines a key trade-off: adding expressive RNN/ bidirectional RNN components improves predictive accuracy but comes at the cost of decreased interpretability in the latent state dynamics. 

GDM identifies three dominant modes with distinct dynamical characteristics, revealed through eigen-analysis and linked to interpretable patterns of social interaction. We illustrate this on an exemplar training session (Figure~\ref{fig:calms}C), where two modes align closely with expert-defined behaviors, while the third (State 3) captures an additional behavioral motif. Importantly, these inferred states do not contradict the annotations; rather, they refine them by decomposing coarse labels into more coherent dynamical patterns. On unseen data, the inferred states can be mapped back to expert labels, while simultaneously providing finer-grained dynamical structure within ambiguous regimes.

\section{Conclusion}
\label{sec:conc}
In this work, we proposed the Gumbel Dynamical Model (GDM), a switching dynamical system defined over relaxed discrete states. By replacing hard switching with Gumbel-driven continuous states, GDM captures smooth, stochastic transitions while preserving interpretable dynamical primitives and enabling efficient gradient-based training. Across synthetic and real-world datasets, GDM improves state stability and alignment with annotations in regimes where discrete-switching models struggle. 

\paragraph{Limitations} Finite-temperature identifiability remains open. In addition, greater model expressivity can reduce interpretability. Our experiments use low- to moderate-dimensional vector observations; end-to-end learning from raw images or videos was not evaluated. A controlled non-Gumbel soft-transition simulator and comparisons with alternative simplex distributions or deterministic softmax also remain untested. Finally, the impact of Gumbel-related parameters on model behavior is not yet fully characterized and warrants further study.

\section*{Acknowledgments and disclosures}
We wish to thank the founder of the Allen Institute, P.G. Allen, for his vision, encouragement, and support. We wish to thank CiCi Zheng, Ian Convy, Adrienne Fairhall, Christof Koch and Karel Svoboda for useful discussions. Y.W. and T.D.K. were supported by the Shanahan Family Foundation Fellowship at the Interface of Data and Neuroscience at the Allen Institute and the University of Washington, supported in part by the Allen Institute.

The authors declare no competing interests.

\bibliography{ref}

\newpage
\appendix

\section{Comparison to Gumbel-Softmax line of work}
\label{app:gumbels}
In this section, we provide detailed comparisons to the prior studies using Gumbel-Softmax in the literature of switching linear dynamical systems in terms of core modeling assumptions, dependency structure, the role of the Gumbel distribution, and inference compatibility. \rev{These approaches retain a separate stochastic continuous trajectory, whereas GDM's relaxed state directly governs observations and remains compatible with standard amortized BBVI.}

\paragraph{Comparison to KVAE} GDM has a fundamentally different graphical structure from the KVAE proposed by \cite{Fraccaro2017}. As illustrated in Figure 1 of both papers, KVAE combines a VAE with a linear Gaussian state space model (LGSSM): observations are first mapped into a low-dimensional latent space by a deep neural network, and those latents are then explained by a soft mixture of LGSSMs. The dynamics-parameter network in KVAE (Section 3.3) corresponds conceptually to the dynamic-operator weighting in p-dLDS \citep{chen2024probabilistic}, where the model learns continuous weights over multiple linear dynamical components.

Crucially, KVAE imposes no structural prior, sparsity constraint, or regularization on these mixture weights, so the learned dynamics tend to be dense mixtures, not interpretable switches. This limitation is directly visible in our experiments: using the publicly released KVAE code, we evaluated the model on the NASCAR benchmark. Table~\ref{tab:nascar} shows consistently low discrete-state inference accuracy for KVAE compared to our model. Moreover, Figure~\ref{fig:nascar}B illustrates an example dynamics-parameter network weighting derived from KVAE, in which the maximum state weights are below 0.30 throughout the time span. 

\rev{The KVAE appendix also discusses approximately discrete dynamics weights using the Concrete distribution, while retaining the LGSSM structure. GDM instead directly governs observations through its stochastic relaxed state.}

\paragraph{Comparison to relaxed SLDS} Although GDM shares the use of the Gumbel distribution with relaxed SLDS proposed by \cite{BeckerEhmck2019}, the resulting generative model is fundamentally different. \rev{\cite{BeckerEhmck2019} use Concrete (Gumbel-Softmax) distributions for the switching prior and posterior while retaining a separate stochastic Gaussian trajectory.} Importantly, GDM directly links states to observations and does not blur latent dynamics with Gaussian noise. \rev{As the authors note below Eq.~7, their observation likelihood depends on the continuous latent trajectory rather than directly on the switching variable.} We further show that removing the important observation-to-state dependency makes GDM equivalent to a three-level mixture SLDS with a deterministic intermediate layer, which is introduced purely for interpretability. Comparing the graphical representations in Figure 1 of both papers highlights these structural differences.

We note that the model of Becker-Ehmck et al. is conceptually similar to the prototype discussed in Appendix~\ref{app:vi3}. We explicitly analyzed its main weakness---the competition between continuous and discrete dynamics to explain the data---and proposed a BBVI-based solution for completeness. We also implemented this variant early in development but found it unsatisfactory --- performing worse than GDM in both accuracy and speed, and scaling poorly to long, high-dimensional time series. Moreover, \rev{they use structured inference for both latent variables}, whereas our approach supports joint sampling and avoids customized inference machinery. 

\rev{Finally, Becker-Ehmck et al. evaluate predictive performance and the information encoded by switching variables. Our focus is on recovering interpretable mixtures of dynamical primitives, with inferred-state accuracy evaluated separately from observation fit.}

\paragraph{Comparison to GS-SNLDS} \rev{Table~\ref{tab:nascar} evaluates SNLDS with collapsed inference, which marginalizes the categorical sequence; GS-SNLDS is a distinct variant discussed here. }The Gumbel–Softmax SNLDS of~\cite{Dong2020} is conceptually and technically distinct from our work, despite superficial similarity. In their method, the Gumbel–Softmax relaxation appears only in the variational posterior as a stand-in for marginalizing discrete states. It does not define the switching dynamics, and the generative model remains a standard SNLDS. In contrast, GDM is a Gumbel-driven dynamical system in which Gumbel noise drives the switching process and determines the latent evolution. Moreover, while amortized inference is standard, the generative structure of GDM makes such amortization fundamentally simpler. In GDM, there is no need to approximate $q(x)$ and no latent-to-latent stochasticity that gradients must pass through. Because observations depend directly on $z$, gradients propagate cleanly through the transition logits, making the model fully compatible with unmodified off-the-shelf amortized inference. 

Additionally, we note that the GS-SNLDS variational posterior is structurally mismatched to its own generative model. It is mean-factorized so that the continuous latent posterior does not depend on the discrete state, even though in SNLDS the discrete state selects the transition dynamics. This breaks a core dependency in the model. This issue is reflected in their experimental results (Table 1): GS-SNLDS performs substantially worse than all other baselines including linear SLDS variants, despite SNLDS being strictly more expressive, empirically demonstrating that the proposed GDM is different in key aspects.

\section{Background}
\label{app:bg}
\paragraph{SLDS} The standard SLDS model generates the observation $y$ from the continuous latent trajectory $x$ and the discrete latent state $z$. The discrete states $z\in\mathbb{R}^K$ can depend on the latent trajectory $x$,
$$z_{t}\sim\mathrm{Cat}(\pi_t),\quad \pi_t = f(z_{t-1}, x_{t-1})$$
where $f$ can be linear or nonlinear. If the discrete state at time $t$ only depends on the latent trajectory at time $t-1$, the model is called recurrent only. 

The continuous latent state $x_t\in\mathbb{R}^D$ follows conditionally linear dynamics determined by state $z_t$,
$$x_{t}\sim\mathcal{N}(A_{z_{t}}x_{t-1}+ b_{z_{t}}, Q_{z_{t}})$$
where $A\in \mathbb{R}^{K\times D\times D}$ are the dynamics matrices, $b\in \mathbb{R}^{K\times D}$ are the shifts, and $Q\in \mathbb{R}^{K\times D\times D}$ are the covariance matrices. $K$ denotes the number of unique discrete states.

Finally, a linear Gaussian observation $y_t\in\mathbb{R}^{N}$ is generated from the corresponding latent state $x_t\in\mathbb{R}^{D}$,
$$y_t\sim\mathcal{N}(Cx_t+d,   \sigma)$$
where $C\in \mathbb{R}^{N\times D}$ is the emission matrix. 

While the variational Laplace expectation-maximization (vLEM) algorithm is a popular choice for inference~\citep{glaser2020recurrent, zoltowski2020general}, it does not guarantee improvement in the evidence lower bound (ELBO) in the E-step because it relies on a second-order Taylor approximation around the mode of the posterior, which can be poor in high-dimensional or multimodal settings.
On the other hand, general stochastic optimization-based variational inference methods such as Black-Box Variational Inference (BBVI) can in principle handle discrete latent variables via score-function estimators. However, in models like SLDS, these estimators typically suffer from high variance when applied to long discrete state sequences, making them difficult to use effectively in practice. In contrast, reparameterization-based gradients cannot be directly applied to discrete latent states, which motivates the use of continuous relaxations such as the Gumbel–Softmax.

\paragraph{BBVI} BBVI uses Monte Carlo gradients to optimize the ELBO. For an SLDS with latent variables $z, x$ and observation $y$,
$$
\mathrm{ELBO} = \E_{q(x, z)}\left(\log p(y, x,z)-\log q_\phi(x, z)\right) \leq \log p_\theta (y)
$$
To optimize the ELBO with stochastic optimization, consider the gradient of the ELBO as an expectation with respect to the variational distribution,
$$\grad_\phi \mathrm{ELBO} = \E_{q(z, x)} \left[\grad_\phi \log q(z, x|\phi)\left(\log p(y, x,z) - \log q(z, x|\phi)\right)\right]$$
Noisy unbiased samples of the ELBO gradient can be computed using Monte Carlo samples from $q(z, x)$.
$$\grad_\phi \mathrm{ELBO} \approx\frac{1}{S}\sum_{s=1}^S \grad_\phi\log q(x_s,z_s|\phi)\left(\log p(y, x_s,z_s) - \log q(x_s, z_s|\phi)\right)$$
Note that the score function and sampling algorithms depend only on the variational distribution, not the underlying model. With samples from the variational distribution, the only requirement is the computation of the log joint $\log p(y, x_s,z_s)$.

\section{Relationship to SLDS}
\label{sec:equiv}
In GDM, the observation $y_t$ at time step $t$ feeds back into the state dynamics through the projection matrix $F$, such that $Fy_t$ recovers the low-dimensional latent trajectory. In fact, GDM can be related to the family of switching linear dynamical systems (SLDS) by introducing a latent projected observation $x_t = \mathbb{E}[F y_{t} \mid z_{\leq t}]$ for $t \geq 1$, where the expectation is taken conditional on all past states. In this section, we prove the equivalence of the 2-level and 3-level formulations and discuss an alternative variational inference method for the 3-level mixture model.

\subsection{Proof of system equivalence}
\label{sec:proof}
In this section, we derive the equivalence relationship between the mixture model and the 2-level GDM. Recall we defined the dependency-removed 2-level GDM as follows,
\begin{align}
\label{eqn:2-level}
z_1&\sim\mathrm{GS}(\pi_1, \tau),\quad z_{t}|z_{t-1}\sim\mathrm{GS}(\pi_t, \tau),\text{ s.t. } \pi_t = f(z_{t-1},\E[Fy_{t-1}|z_{\leq t-1}]),\quad t\geq 2 \\ \nonumber
y_1&|z_1\sim \mathcal{N}(\sum_k z_{1,k}\mu_k, R_t),\quad y_t|y_{t-1}, z_t\sim\mathcal{N}(\sum_k z_{t,k} (S_{k}Fy_{t-1}+ b_{k}), R_t),\quad t\geq 2 
\end{align}

And we defined the 3-level mixture model as follows (see eqn. (\ref{eqn:3-level})),
\begin{align}
z_1&\sim\mathrm{GS}(\pi_1, \tau), \quad z_{t}|z_{t-1},x_{t-1}\sim\mathrm{GS}(\pi_t, \tau),\text{ s.t. }\pi_t = f(z_{t-1},x_{t-1}),\quad t\geq 2 \label{eqn:z-level}\\ 
x_1&=\sum_k z_{1,k}\mu_k, \quad x_{t}| x_{t-1}, z_t=\sum_k z_{t,k} (A_{k}x_{t-1}+ c_{k}), \quad t\geq 2\label{eqn:x-level}\\  
y_t&|x_t\sim\mathcal{N}(Cx_t, Q_t) ,\quad t\geq 1\label{eqn:y-level}    
\end{align}

Firstly, we derive the 3-level mixture model (\ref{eqn:3-level}) from system \ref{eqn:2-level}. The state transition equation of model (\ref{eqn:3-level}) follows from a straightforward substitution. To obtain eqn.(\ref{eqn:x-level}), we consider
$$\E_{y_t|z_{\leq t}}(Fy_{t}| z_{\leq t})=F\E_{y_{t-1}| z_{\leq t}}[\E_{y_t|y_{t-1}}(y_t|y_{t-1},z_{\leq t})]$$
Splitting time steps before $t$ into time steps before $t-1$ and time step $t$ we have, 
\begin{align*}
\E_{y_t|z_{\leq t-1}, z_t}(Fy_{t}|z_{t-1}, z_t)&= \E_{y_{t-1}|z_{\leq t-1}, z_t}\sum_k z_{t,k} F(S_{k}(Fy_{t-1})+ b_{k})\\
&=\sum_k z_{t,k}\E_{y_{t-1}|z_{\leq t-1},z_t}(FS_{k}(Fy_{t-1})+ Fb_{k})\\
&=\sum_k z_{t,k}(FS_{k}x_{t-1}+ Fb_{k})
\end{align*}
The last line is derived from the definition $x_{t-1} = \E(Fy_{t-1}|z_{\leq t-1})$ and the fact that $y_{t-1}$ and $z_t$ are conditionally independent given $z_{t-1}$. Conditioning on $x_{t-1}$ and $z_t$, $x_t=\E_{y_t|z_{\leq t}}(Fy_{t}|z_{\leq t})$ is equivalent to the LHS of eqn.(\ref{eqn:x-level}), as $x_{t-1}$ is fully determined by states before time step $t-1$. The RHS of the equation above can be put into RHS of eqn.(\ref{eqn:x-level}) by setting $A_k = FS_k$, and $c_k = Fb_k$. Finally, to obtain eqn.(\ref{eqn:y-level}), we consider the mean and variance of $y_t$. If we set $C = F^\dagger$, we have $\E(y_{t}|x_t) = Cx_t$. To obtain the variance, we consider
\begin{align*}
Q_t = \mathrm{Var}(y_t|x_t) &= \E\mathrm{Var}(y_t|y_{t-1},x_t, z_t) +\mathrm{Var}\E(y_t|y_{t-1}, x_t, z_t)\\
& = R_t+\mathrm{Var}(\sum_k z_{t,k} (S_{k}Fy_{t-1}+ b_{k}))
\end{align*}
We can remove the dependency on $x_t$ in both summation terms, since $x_t$ is fixed given $z_t$ and $z_{t-1}$, and $y_t$ is independent of $z_{t-1}$ given $z_t$. In practice, we can assume a diagonal covariance structure $R_t=\sigma I$.

Next, we show the reverse derivation from the mixture model to the GDM. 

To obtain the Gumbel dynamics equation for the GDM, we consider
$$\E_{y_t|z_{\leq t}}(Fy_{t}|z_{\leq t})= \E_{x_t| z_{\leq t}}\E_{y_t|x_t,z_{\leq t}}(Fy_{t}|x_t,z_{\leq t}) =\E_{y_t|x_t}(Fy_{t}|x_t) $$
The inner expectation reduces to $\E_{y_t|x_t}(Fy_{t}|x_t)$ as $y_t$ is independent of $z_t$ given $x_t$. The outer expectation can be removed as $x_t$ is fully determined by states before time step $t$. 

By eqn. (\ref{eqn:y-level}), we know that 
$$x_t = \E_{y_t|x_t}(Fy_{t}|x_t)= \E_{y_t|z_{\leq t}}(Fy_{t}|z_{\leq t}) $$
where $F=C^\dagger$. This gives the Gumbel dynamics equation for the GDM by substituting $\E(Fy_{t-1}|z_{\leq t-1})$ in eqn. (\ref{eqn:z-level}).

To derive the observation level for the GDM, we substitute eqn. (\ref{eqn:x-level}) into eqn. (\ref{eqn:y-level}). Specifically, we write $y_t = Cx_t + \epsilon$ where $\epsilon \sim\mathrm{N}(0, Q)$. Then we have,
\begin{align*}
y_t &= C\sum_k z_{t,k} (A_{k}x_{t-1}+ c_{k}) + \epsilon\\
& = C\sum_k z_{t,k} (A_{k}(Fy_{t-1}-\tilde{\epsilon})+ c_{k}) + \epsilon\\
& =  \sum_k z_{t,k} (CA_{k}Fy_{t-1}+ Cc_{k}) + \epsilon - \sum_k z_{t,k}CA_k\tilde{\epsilon}
\end{align*}
where $\tilde{\epsilon}\sim\mathcal{N}(0,FQF^\intercal)$ is another Gaussian noise term. The second line comes from eqn. (\ref{eqn:y-level}), as we have $Fy_{t-1} = x_{t-1} +\tilde{\epsilon}$ where $\tilde{\epsilon}\sim\mathcal{N}(0,FQF^\intercal)$. Therefore, if we set $S_k = CA_k$, $b_k = Cc_k$ and $R_t = Q+ (\sum_k z_{t,k} CA_k)FQF^\intercal (\sum_j z_{t,j}A_j^\intercal C^\intercal$), we recover the observation dynamics in GDM. 

We note that this substitution introduces a composite noise term that depends on both current and previous Gaussian perturbations, yielding an effective moving-average structure in the observation noise. As a result, the covariance $R_t$ becomes time-dependent even when $Q$ is diagonal. In practice, however, we typically assume time-homogeneous and conditionally independent Gaussian noise (i.e., $R_t=R$ and $Q_t=Q$) when fitting models. Therefore, the equivalence is exact when allowing general, potentially time- and state-dependent covariance structures. Under the common restriction to conditionally independent, time-homogeneous noise, the two formulations remain equivalent in their deterministic dynamics, while differing in their induced noise processes.

\subsection{Variational inference for 3-level mixture model}
\label{app:vi3}
As discussed in the main text, inference for the general 3-level mixture model is more challenging as we need to define variational distributions for both the latent variables $x$ and $z$. We define a joint variational distribution $q(x_{1:T}, z_{1:T})$ that captures dependencies between $x$ and $z$, implemented through an alternating sampling scheme.  For $z$, we define a conditional variational distribution that depends on the current estimate of $x$, i.e., $q(z_{1:T}|x_{1:T}) = q(z_1)\prod_{t=2}^Tq(z_t|z_{t-1}, x_{t-1})$. For $x$, we introduce dependencies that span multiple time steps by assuming a Gaussian with block tri-diagonal precision for $x_{1:T}$. 
$$q(x_{1:T}|z_{1:T}) = \mathcal{N}(x_{1:T}|\mu, \Sigma) = \mathcal{N}(x_{1:T}|J, h) $$
where $J$ is the precision matrix, $h$ is the linear potential, $\mu=J^{-1}h$ is the mean, and $\Sigma = J^{-1}$ is the inverse precision (covariance) matrix. It can be written as the following pairwise linear Gaussian dynamics,
$$q(x_{1:T})
\propto [\prod_{t=1}^{T-1} \mathcal{N}(x_{t+1} | A_t x_t + b_t, Q_t)]
\cdot [\prod_{t=1}^T \mathcal{N}(x_t | m_t, R_t)]$$
Note that it is easier to work with the pairwise LDS structure as the precision matrix $J$ can be efficiently inverted and sampled from. We assume that the transition parameters $A_t$, $Q_t$, and $b_t$ are state-dependent, $A_t =A_{z_t} $, $b_t = b_{z_t}$, and $Q_t =Q_{z_t}$. Note that the parameters of this distribution depend on the sampled trajectory of $z$, inducing an implicit coupling between $x$ and $z$.

\paragraph{Sampling mechanism} Note that sequential sampling is feasible for $z$ but not for $x$. Recall the standard way of sampling from $\mathcal{N}(\mu,\Sigma)$ as follows. If $\Sigma$ has Cholesky decomposition $\Sigma= LL^\intercal$, then we can generate samples using $x = \mu + L\eta$ where $\eta \sim \mathcal{N}(0,I)$.
In our case, we need $z_t$ for all time steps $t$ to compute linear potential $h$ and precision matrix $J$. To sample from $J$, we solve two equations: $J\mu = h$ and $U^\intercal \tilde{x} = \eta$ where $U$ is the Cholesky decomposition of $J$ s.t. $J = UU^\intercal$. The final sample of $x$ is the sum of $\mu$ and $\tilde{x}$. 

To sample from $q(x,z)$, we use an alternating procedure. We first initialize $x_{1:T}$ from the observations $y_{1:T}$. Given $x$, we sample $z$ sequentially as follows:  1) Sample $z_1$ from the GS distribution with $\phi_1$; 2) Compute logits $\phi_t$ using the learnable transition function and sample $z_t$, for all $t\geq 2$. Given the sampled $z_{1:T}$,  we then construct the Gaussian parameters $J$ and $h$ and sample $x_{1:T}$ from $q(x_{1:T}|z_{1:T})$ using the linear-Gaussian structure described above. Note that this alternating scheme defines a valid joint sampler without requiring an explicit factorization of $q(x,z)$ into mutually conditioning components.

\paragraph{Complete ELBO}

The ELBO for the 3-level mixture model is: 
\begin{align*}
\log p_\theta(\boldsymbol{y}_{1:T}) \geq &\E_{q(x,z)}\left[\log p(y,x,z) - \log q(x,z)\right]\\
= &\mathbb{E}_{q(x,z)} \left[ \sum_{t=1}^{T} \log p(y_t | x_t) + \log p(x_1 | z_1) + \sum_{t=2}^{T} \log p(x_t | x_{t-1}, z_t) \right.\\
&\quad\left. + \log p(z_1) + \sum_{t=2}^{T} \log p(z_t | z_{t-1},x_{t-1}) \right]\\
& - \mathbb{E}_{q(x, z)} \left[  \log q(x_{1:T}|z_{1:T}) + \log q(z_1) + \sum_{t=2}^{T} \log q(z_t | z_{t-1}, x_{t-1}) \right]
\end{align*}

\section{Identifiability considerations for GDM}
\label{app:identifiability}
In the limiting case $\tau\to 0$, the GDM has an equivalent formulation as a finite mixture model analogous to an AR‑HMM. Following the same notations in ~\cite{BalsellsRodas2023}, for finite horizon $T$, one can define a bijective path indexing function $\psi$ that maps each $i\in\{1,\ldots, K^T\}$ to a set of states $z_{1:T}$. Then, the family of GDMs can be seen as a finite mixture over all possible discrete state paths.

Let $M_k =S_kF$ denote the product of the dynamic matrix $S_k$ and projection matrix $F$.  At the observation level, GDM then satisfies the \textit{unique-indexing assumption} on Gaussian means and initial states used in \cite{BalsellsRodas2023}. By Theorem 3.2 in their paper, under these conditions, the family of GDMs is identifiable up to permutations. Importantly, this identifiability result does not require restricting the form of the state transitions, and arbitrary recurrence from the switches is allowed~\cite{rodas2021casual}. In the case that transition logits $\pi_t$ depend only on the previous discrete state $z_{t-1}$, one can uniquely recover the transition matrix.

For $\tau>0$, identifiability becomes more subtle. The continuous GS relaxation means that latent state $z_t$ at each time step $t$ takes values on the simplex $\Delta^{K-1}$, so GDM is no longer a finite mixture but rather behaves like an infinite mixture over continuous paths. The considerations above rely heavily on the use of finite mixture modeling techniques, and characterizing identifiability in the non-limiting regime remains an open problem. \rev{The categorical-limit argument therefore does not provide an identifiability theorem for finite-temperature GDM; interpretation in that regime rests on the learned primitives and empirical state-recovery controls.}

Developing a full identifiability theory for the non-limiting case will require new mathematical statements. A potential route toward a formal proof may draw on the ideas in~\cite{Barndorff1965}.

\section{Computational advantages and setups}
\label{app:setup}

\rev{We used official baseline implementations and their recommended training and test-time inference procedures. To keep observation-model flexibility comparable, we simplified the observation layers of SNLDS and KVAE to match those of the linear models. Hyperparameters, including learning rate and latent dimension, were selected per dataset using the appropriate validation objective on held-out training trials or segments; neither test data nor expert labels were used for selection. We held $K$ equal across methods in direct state-recovery comparisons.}

GDM replaces variational EM with fully differentiable amortized inference, leading to two computational advantages:
\begin{itemize}
\item \textbf{Training time} Each iteration of GDM uses standard backpropagation with reparameterized Gumbel-Softmax samples, avoiding the iterative E-step optimization over latent trajectories required in prior methods. Empirically, this yields substantially faster wall-clock convergence under comparable architectures and hardware (as discussed below).
\item \textbf{Inference} Latent states are obtained via a single forward pass, whereas baselines often require per-sequence optimization over latent trajectories, leading to substantially higher test-time cost (often an order-of-magnitude difference in practice).
\end{itemize}
To isolate per-iteration efficiency, Table~\ref{tab:runtime} reports per-iteration wall-clock time across methods and datasets. Experiments were conducted on a single GPU (NVIDIA g4dn.4xlarge instance, 16GB GPU RAM). Batch sizes were matched across methods. As shown in Table~\ref{tab:runtime}, GDM consistently achieves lower cost, especially as sequence length and dataset size grow. 
\begin{table}[h!]
\centering
\begin{tabular}{lccccc}
	\toprule
	& SLDS & rSLDS & pdLDS & SNLDS & GDM \\
	\midrule
	Nascar & 0.080 & 0.130 & 0.920 & 2.12 & 0.010 \\
	F1     & 1.650 & 3.860 & 20.25 & \textemdash & 0.036 \\
	CalMS  & \textemdash & 54.88 & \textemdash & \textemdash & 1.18 \\
	\bottomrule
\end{tabular}
\caption{Per-iteration wall-clock time (in seconds) across datasets. Lower values indicate better performance. Missing entries correspond to CUDA out-of-memory errors or execution timeouts. }
\label{tab:runtime}
\end{table}

In addition to computational efficiency, inference strategies differ qualitatively across models and impact generalization behavior. As discussed in the main text, most SDS baselines require re-optimizing variational posteriors on new sequences to achieve strong test performance, reflecting the lack of amortization. In contrast, GDM performs inference via a single forward pass.

KVAE and SNLDS also support amortized inference and can generalize without retraining, but rely on additional architectural or inference design choices. KVAE introduces two separate networks: a VAE that maps observations into a low-dimensional latent trajectory, and a dynamics-parameter network that maps this trajectory to time-varying system parameters. SNLDS, which inherits collapsed variational inference, supports expressive latent dynamics and amortized inference for continuous variables; however, because discrete switching variables are marginalized out rather than inferred directly, the discrete states must be recovered post hoc, limiting interpretability.

Combined with the absence of test-time optimization, this leads to significantly faster end-to-end performance for GDM.

\section{More discussions on the NASCAR dataset}
\label{app:nascar}
The full generative model used to simulate the NASCAR dataset is described as follows,
\begin{align}
z_1&\sim\mathrm{GS}(\pi_1, \tau), \quad z_{t}|x_{t-1}\sim\mathrm{GS}(Tx_{t-1}+r, \tau)\quad t\geq 2 \label{eqn:trans}\\ 
x_1&=\sum_{k=1}^{4} z_{1,k}\mu_k, \quad x_{t}| x_{t-1}, z_t=\sum_{k=1}^{4} z_{t,k} (A_{k}x_{t-1}+ c_{k}) \quad t\geq 2 \label{eqn:dyn}\\  
y_t&|x_t\sim\mathcal{N}(Cx_t, \sigma I) ,\quad t\geq 1   
\end{align}

This can be achieved by setting extreme Gumbel-Softmax logits in eqn. (\ref{eqn:trans}). As an example, the transition matrix $T$ and the bias $r$ can be defined as
\[
T =
\begin{bmatrix}
10 & 0 \\
-10 & 0 \\
0 & 10 \\
0 & -10
\end{bmatrix}
\qquad
r =
\begin{bmatrix}
-20 \\
-20 \\
-10 \\
-10
\end{bmatrix}
\]
Eqn. (\ref{eqn:trans}) can be viewed as a classifier that divides the space into four regions such that the logit of each region $k$ is computed as $T_k \cdot x + r_k$ where $x\in\mathbb{R}^2$ denotes the point on the 2D trajectory. For example, if $x_1>2$ and $-1<x_2<1$, the first logit will be greater than $0$ while other logits will be smaller than $0$, so the point is highly likely to be classified in state $k=1$.

Eqn. (\ref{eqn:dyn}) specifies how the system moves in each state. For the standard NASCAR, 
the ground truth dynamics matrices are defined as,
\[
A_1 =A_2 =\mathrm{expm}\left(
\begin{bmatrix}
0 & \frac{\pi}{24} \\
-\frac{\pi}{24} & 0
\end{bmatrix}
\right),
\quad
A_3 =A_4 =I=
\begin{bmatrix}
1 & 0 \\
0 & 1
\end{bmatrix}
\]
where the first two states correspond to the semicircular turns of $7.5^\circ$ at the end of the straight track. The ground-truth offsets are defined as,
\[
c_k =
\left\{
\begin{array}{ll}
-(A_1 - I) \cdot \mathrm{FP}_1, & k = 1 \\
-(A_2 - I) \cdot \mathrm{FP}_2, & k = 2\\
\begin{bmatrix} 0.1& 0 \end{bmatrix}, & k = 3 \\
\begin{bmatrix} -0.25 & 0 \end{bmatrix}, & k = 4 \\
\end{array}
\right.
\]
where $c_1$ and $c_2$ specify rotations around $\mathrm{FP}_1=(2,0)$ and $\mathrm{FP}_2=(-2,0)$ at the semicircular turns, while $c_3$ and $c_4$ specify the constant speed along the straight track. 

To model variable-speed transitions, we may introduce another parameter $s$ that denotes a varied speed for the dynamics equation (\ref{eqn:dyn}) such that $\tilde{c}_k = sc_k$ where $s\in[s_{min},1]$ is uniformly sampled between a minimum low speed $s_{min}$ and full speed and is applied throughout the linear segment of the track. The observation is generated in the same way as before. Given the previous location in the trajectory $x_{t-1}$ and the current state $z_t$, we can generate the next trajectory point using eqn. (\ref{eqn:dyn}). The trajectory is then mapped to the observations. Note that the shape of the trajectory will not be changed fundamentally by varying speed as the movement direction of each state remains unchanged.

\section{Additional results for F1 dataset} \label{app:usage}

\begin{figure}[!htbp]
\centering
\includegraphics[width=\textwidth]{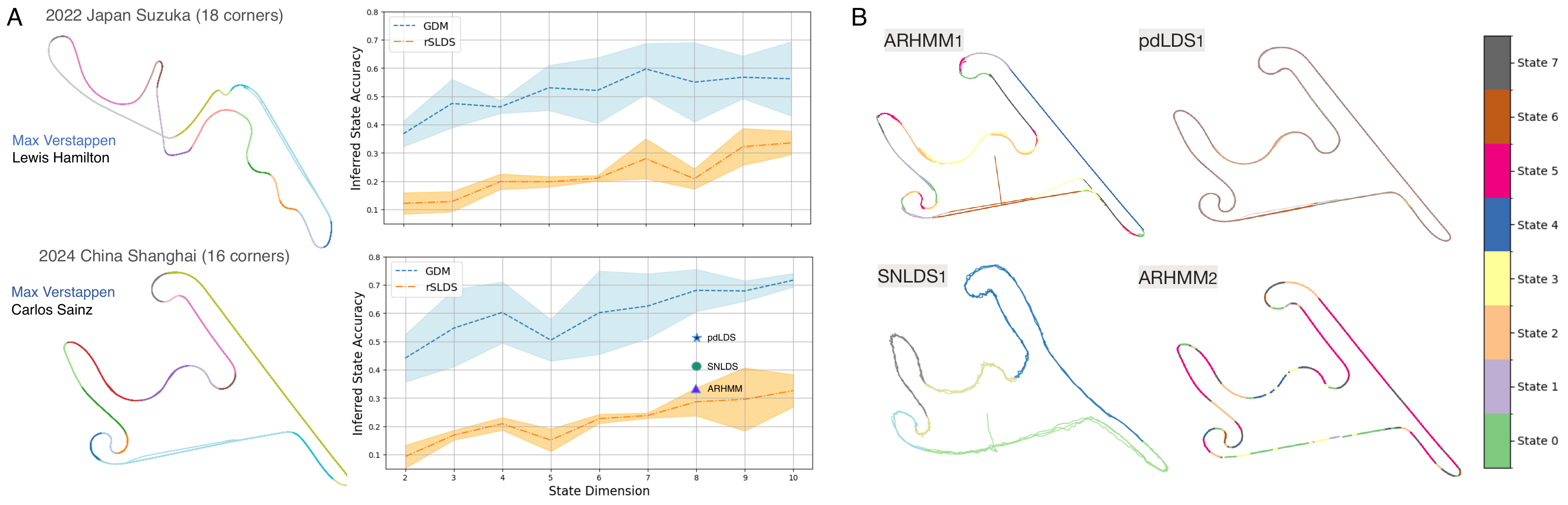}
\caption{A. Comparison of inferred state accuracy between GDM and rSLDS across state dimensionalities on the Shanghai International Circuit and Suzuka Circuit. Performance is evaluated over 5 train/test splits with different random seeds; shaded regions denote standard deviation across seeds. GDM consistently achieves higher state accuracy, particularly at low dimensions. B. Example inferred trajectories from additional baselines not shown in the main figure, including ARHMM, pdLDS, and SNLDS for the position-based setting, and ARHMM for the velocity-based setting. The corresponding inferred state accuracy for this single run is marked in Figure~\ref{fig:extra}A. }
\label{fig:extra}
\end{figure}

\begin{figure}[!htbp]
\centering
\includegraphics[width=\textwidth]{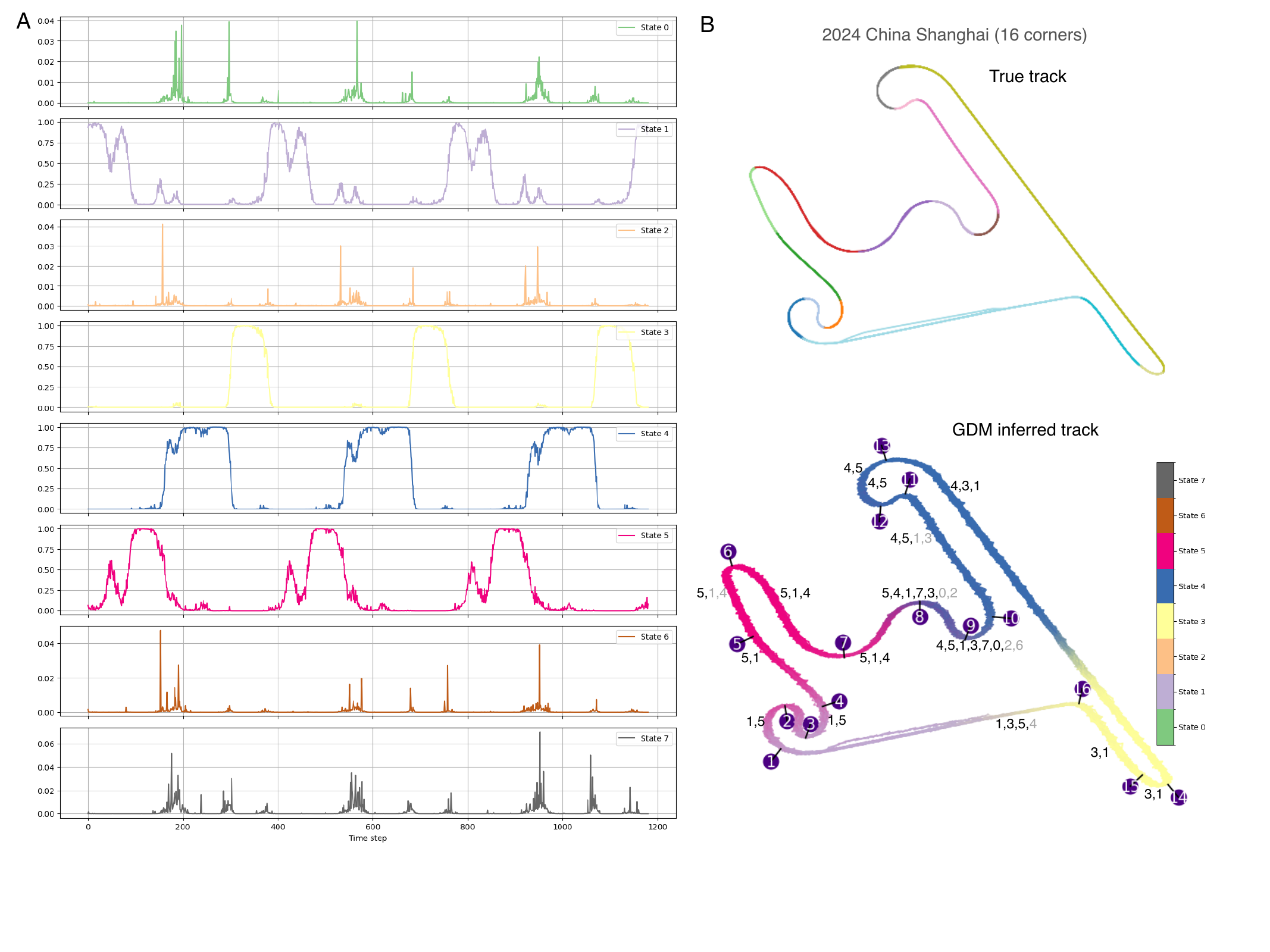}
\caption{A. Complete state usages for Figure~\ref{fig:f1}C. Example inferred trajectory for GDM, with complementary states annotated in grey. }
\label{fig:usage}
\end{figure}

As mentioned in the main text, GDM utilizes all states, but not equally. In Figure~\ref{fig:usage}A, we show the complete state usage of GDM for the trial illustrated in Figure~\ref{fig:f1}C. For demonstration purposes, we display the first three laps around the track. As seen in the plot, while all states capture the three laps as three clear peaks in probability, States 1, 3, 4, and 5 are more dominant than the other four states. This is also reflected in the state annotations in Figure~\ref{fig:f1}C. Here, we provide a more detailed version of Figure~\ref{fig:f1}C by lowering the presence threshold to 5\% of all time steps associated with the expert-labeled state. The complementary states for each segment are greyed out.

The unequal usage of states helps explain the observation that the inferred state accuracy of GDM improves rapidly in the initial steps and then plateaus. GDM allocates additional states to less dominant roles, so the marginal gain of increasing the number of states decreases after the first few.

For practical visualization, we put an emphasis on the dominant states. Specifically, we set transparency to the maximum value of state proportions at each time step and mix colors according to the proportions of active states. This yields a gradual change in color across transitions and more transparent segments where mixtures of overlapping states occur.

\section{Ethics statement}
\label{app:ethics}
We conceived GDM as a tool to improve analysis of dynamical phenomena. While we hope that it will benefit society in the longer run by supporting progress across scientific disciplines, we do not think our work carries any immediate societal impact.


\end{document}